\documentclass[twocolumn,10pt]{article}
\usepackage[left=0.5in,right=0.5in,top=0.7in,bottom=1.0in]{geometry}

\usepackage{amssymb}
\usepackage{amsmath}
\usepackage{booktabs}
\usepackage{makecell}
\usepackage{csvsimple}
\usepackage{siunitx}
\usepackage[ruled]{algorithm2e}
\usepackage{listings}
\usepackage{pgfplotstable}
\usepackage{multicol}

\usepackage{graphicx}
\usepackage[citations]{markdown}
\graphicspath{{figures/}}

\usepackage{tikz}
\usepackage{tkz-euclide}
\usetikzlibrary{calc}
\usetikzlibrary{math}
\usetikzlibrary{shapes}
\usetikzlibrary{decorations.pathreplacing}

\usepackage[backend=biber,style=authoryear,uniquename=false,uniquelist=false,doi=false,url=false,maxcitenames=2]{biblatex}
\bibliography{scanpaths}

\usepackage{enumitem}
\setlistdepth{20}
\renewlist{compactitem}{itemize}{20}
\setlist[compactitem]{label=$\bullet$,nosep}

\makeatletter
\newcommand{\smallparagraph}{%
  \@startsection{paragraph}{4}%
  {\z@}{0.5ex \@plus 0.2ex \@minus .2ex}{-1em}%
  {\normalfont\normalsize\bfseries}%
}
\makeatother

\begin{document}


\title{State-of-the-Art in Human Scanpath Prediction}
\author{Matthias Kümmerer \and Matthias Bethge}
\date{%
    {\texttt{matthias.\{kuemmerer,bethge\}@bethgelab.org}}\\[2.5ex]%
    University of Tübingen%
}
\maketitle

\abstract{
    The last years have seen a surge in models predicting the scanpaths of fixations made by humans when viewing images.
    However, the field is lacking a principled comparison of those models with respect to their predictive power.
    In the past, models have usually been evaluated based on comparing human scanpaths to scanpaths generated from the model.
    Here, instead we evaluate models based on how well they predict each fixation in a scanpath given the previous scanpath history.
    This makes model evaluation closely aligned with the biological processes thought to underly scanpath generation and allows to apply established saliency metrics like AUC and NSS in an intuitive and interpretable way.
    We evaluate many existing models of scanpath prediction on the datasets MIT1003, MIT300, CAT2000 train and CAT200 test, for the first time giving a detailed picture of the current state of the art of human scanpath prediction.
    We also show that the discussed method of model benchmarking allows for more detailed analyses leading to interesting insights about where and when models fail to predict human behaviour.
    The MIT/Tuebingen Saliency Benchmark will implement the evaluation of scanpath models as detailed here, allowing researchers to score their models on the established benchmark datasets MIT300 and CAT2000.
}



\begin{figure}[t]
    \begin{center}
        \includegraphics[width=0.9\linewidth]{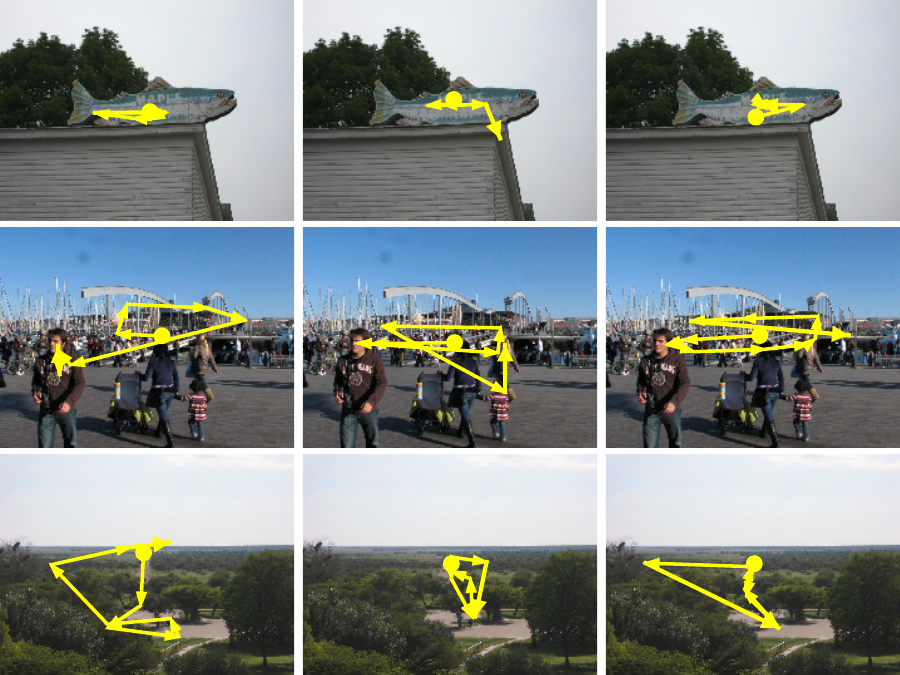}
    \end{center}
    \caption{Humans inspect images in scanpaths consisting of fixations. Here we show scanpaths made by different subjects for three different images. The yellow dot shows the initial central fixation, the yellow arrows connect the consecutive fixations in the scanpath.}
    \label{fig:scanpaths}
\end{figure}

\begin{figure*}[t]
    \begin{center}
        \includegraphics[width=0.9\textwidth]{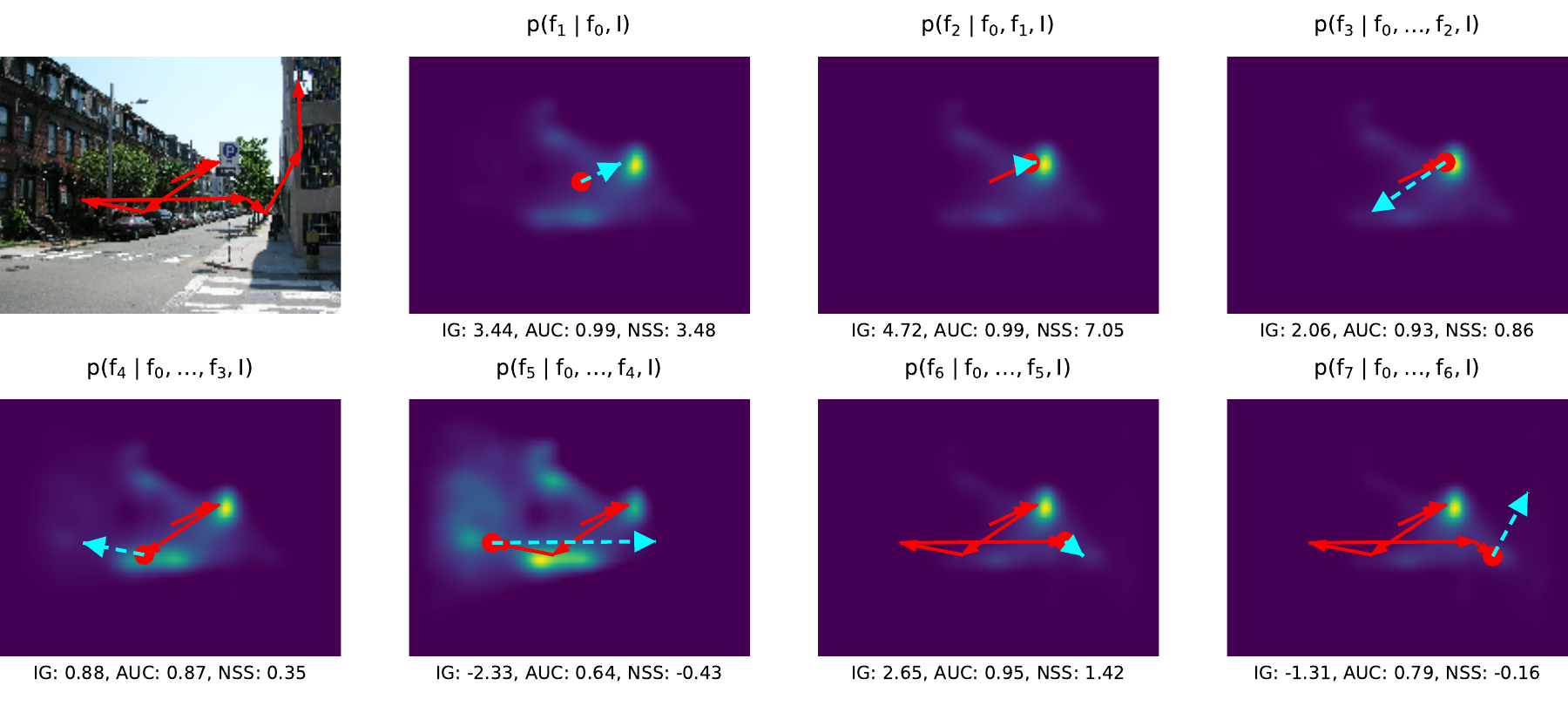}
    \end{center}
    \caption{Conditional predictions. Top left: an example image and a ground truth scanpath. Other panels: for each fixation $f_i$ the model is provided with the image $I$ and the previous fixations $f_0, \dots, f_{i-1}$ (red arrows) and produces a conditional prediction $p(f_i \mid f_0, \dots, f_{i-1}, I)$ (displayed heatmap). Each such conditional prediction can then be evaluated with respect to how well it predicts the correct next fixation location (end of cyan arrow).}
    \label{fig:conditional_prediction}
\end{figure*}

\section{Introduction}

Humans gather visual information about their environment using their eyes.
Due to anatomic constrains, we can collect high-density information only for a small central area of our field of view.
Therefore, we move our eyes, directing them towards whatever might be most relevant at the moment.
How we choose where to look as attracted a lot of research over the decades: eye movements are an overt manifestation of the decision processes underlying attention.
In particular, the question which image features affect the spatial distribution of fixations has been the focus of a large amount of research.
Starting with the seminal model of Itti and Koch \parencite{ittiModelSaliencybasedVisual1998}, computational models of \textit{saliency prediction} tried to predict so called \textit{saliency maps} capturing which parts of an image observers might look at most.
The field of saliency models is now a very mature field with many new models being published every year and with established benchmarks tracking the progress and state-of-the-art \parencite{borjiStateoftheArtVisualAttention2013}.
However, saliency map models focus purely on the spatial distribution of fixations.
But fixations are not independent of each other:
we explore an image in a \textit{scanpath} of fixations and where we looked earlier can influence later fixations in a scanpath.
So called \textit{scanpath models} try to take this into account by proposing mechanisms generating complete scanpaths.
While this field is not as active as the field of saliency modelling, the last years have seen the rise of many models of scanpath prediction, some implementing hypothesized biological or cognitive process and some other models purely being trained for maximum prediction performance.

Although by now there is a substantial number of models of scanpath prediction, there is no clear picture of the state of the field.
One reason for this might be that the field is not only lacking clearly established benchmarks as, e.g., provided by the MIT/Tuebingen Saliency Benchmark \parencite{juddBenchmarkComputationalModels2012,kummererSaliencyBenchmarkingMade2018} for the saliency community.
Already the question how to judge the predictive power of a scanpath model doesn't have an agreed-upon answer.
Researchers usually compare scanpaths generated by a model to human scanpaths, either simply by visual inspection or using one of many sequence metrics that again have many hyperparameters concerning scanpath simplification, scanpath alignment, gap penalties and fixation distance metrics.

Here, we choose a different approach.
Instead of comparing model samples to real data, we evaluate how well models predict the real scanpaths, fixation by fixation, using \textit{conditional} (scanpath history dependent) \textit{priority maps}.
This approach is very closely aligned with the assumed biological generative process underlying scanpaths.
Additionally, it takes into account the full complexity of model predictions instead of collapsing the prediction into potentially very few samples.
The approach is a straight forward extension of how saliency benchmarking has been done in the past for spatial saliency prediction and allows to rank spatial models and scanpath models on the same well established and understood metrics and dataset in the same benchmark leaderboards.
This allows us, for the first time, to give a detailed overview of modeling performance in the field of human scanpath modeling.

Our contributions in this paper are as follows:
\begin{itemize}
    \item We establish a principled framework for evaluating models of human scanpath prediction
    \item We summarize the state of the field
    \item We provide a overview of how well many different models perform on different datasets and pinpoint common sources of suboptimal predictions.
    \item We establish a public benchmark where researchers can submit their future models to see how well the do compared to previous models.
\end{itemize}

\section{Scanpath Prediction and Benchmarking}
\label{sec:scanpath_prediction}

We don't expect the human brain to first generate a full scanpath and then execute it fixation by fixation.
Instead, results from Neuroscience suggest \parencite{kalesnykasPrimateSuperiorColliculus1996,girardBrainstemCortexComputational2005,krauzlisSuperiorColliculusVisual2013} that while the eye fixates an image location, the brain selects the next point to look at given the current fixation position, the internal brain state which can incorporate memory, task and biases and other factors such as occulomotor constraints and then executes the next saccade after which the process repeats.
Virtually all scanpath models indeed implement this process: they generate fixations in an iterative process where the current and previous fixation positions influence where the next fixation is chosen.

\begin{algorithm}[t!]
    \DontPrintSemicolon
    \KwData{image, N}
    \KwResult{scanpath}
    scanpath = [screen\_center]\;
    state = initialize(image, screen\_center)\;
    i = 1\;
    \While{i $<$ N}{
        priority\_map = compute\_priority\_map(state)\;
        next\_fixation = sample\_fixation(priority\_map)\;
        scanpath = scanpath + [next\_fixation]\;
        state = update\_state(state, next\_fixation)\;
        i = i + 1\;
    }
    \caption{Sampling Model}
    \label{alg:sampling}
\end{algorithm}

However, when it comes to evaluating the predictive quality of scanpath models on ground truth data, all commonly used approaches ignore this underlying generative process: A typical approach is to generate new scanpaths from the model and then use some kind of distance function to compare generated scanpaths from human scanpaths.
Here, we suggest a method that more closely follows 
how we think scanpaths are generated:
\textit{For each fixation in each scanpath}, we evaluate individually how well it is predicted by the model when informing the model about the previous fixations that the subject made in the image.
Essentially, we treat each fixation in a scanpath as a decision to move the eye from the current fixation location to a new fixation location and we asses how well a model predicts this decision.

Figure \ref{fig:conditional_prediction} demonstrates this evaluation method: The upper left (a) shows an example image with an example human ground truth scanpath.
Next (b), we tell a model that the initial (forced) fixation of the subject was in the center of the image and ask it for a prediction about the next fixation.
This is a simple spatial fixation prediction task.
If the model predicts some kind of priority map or fixation distribution we can evaluate it using many of the established metrics from the field of saliency prediction, such as AUC, NSS or (in the case of probabilistic models) log-likelihoods and information gain.
This process can then be repeated for each fixation in the scanpath:
The model is informed about the previous scanpath and computes a prediction for where it expects the next fixation to be placed (c)-(j).
To compute overall model scores, the scores for fixations can be averaged.

In the case of probabilistic models, this corresponds to splitting the joint probability distribution over scanpaths $p(f_0, \dots, f_n)$ via the chain rule into a product of conditional distributions $\prod_i^n p(f_i \mid f_0, \dots, f_{i-1})$.

Evaluating scanpath prediction performance fixation by fixation has several advantages besides being more closely aligned with our assumption of the underlying process in the brain.
Firstly, the full information in the model predictions is used for evaluation.
This sets it apart from all sampling-based metrics which reduce the model predictions to a single fixation which ignores a lot of information such as multi-modality.
Secondly, evaluating fixation positions in conditional model predictions makes the evaluation compatible with the evaluation of classic spatial fixation prediction models, allowing to compare, e.g., the effect of better spatial saliency modeling and the effect of better incorporation of oculumotor biases.
Thirdly, as opposed to the many hyperparameters of most scanpath-sequence-metrics, this approach has no hyperparameters at all.
Fourthly, assigning scores to individual fixations instead of whole scanpaths allows for much more detailed analyses of which fixations cause problems in modeling.
We will demonstrate this later in more detail (see Section \ref{sec:results_case_studies}).
Finally, our approach avoids evaluation artifacts as, e.g., in multi-modal distributions.
Multi-modal distributions are expected to be common in scanpath distributions, e.g., when there are multiple interesting objects in an image.
In this case, evaluating samples can result in a performance as extreme as 0\% or 100\%, depending on whether the evaluated sample comes from the same mode of the distribution or not
and early divergence from the reference scanpath has an unproportionally large influence on the score
Conditional priority can clearly show predicted multi-modal structures in visual inspection and lead to more intuitive performance scoring.

\begin{algorithm}[t!]
    \DontPrintSemicolon
    \KwData{image, scanpath\_history}
    \KwResult{priority\_map}
    // \textit{scanpath = [scanpath\_history[0]]}\;
    state = initialize(image, scanpath\_history[0])\;
    i = 1;\quad N=length(scanpath\_history)\;
    \While{i $<$ N}{
        // \textit{priority\_map = compute\_priority\_map(state)}\;
        next\_fixation = scanpath\_history[i]\;
        // \textit{scanpath = scanpath + [next\_fixation]}\;
        state = update\_state(state, next\_fixation)\;
        i = i + 1
    }
    priority\_map = compute\_priority\_map(state)\;
    \caption{Computing conditional priority maps}
    \label{alg:conditionalprediction}
\end{algorithm}

This approach of evaluating scanpaths has been rarely used in the past.
It was most often used in the context of probabilistic models, where \textcite{kummererInformationtheoreticModelComparison2015},  \textcite{schuttLikelihoodbasedParameterEstimation2017}, \textcite{clarkeSaccadicFlowBaseline2017} and \textcite{kummererDeepGazeIIIUsing2019} evaluated log likelihoods per fixation for human scanpaths.
\textcite{barthelmeModelingFixationLocations2013} also computed log likelihoods, but they applied the framework of point processes to score whole scanpaths instead of individual fixations.
For classic saliency metrics, \textcite{tatlerLATESTModelSaccadic2017} evaluated AUC scores on what we call conditional priority map here.
However, so far these approaches were not unified to compare both probabilistic and nonprobabilistic models.
There is no comparative account of a larger number of models of scanpath prediction in terms of their prediction performance.
This is what we present in this work.

\subsection{Application to existing models}
\label{sec:modifying_models}


The proposed method of evaluating conditional fixation predictions can be readily applied to most existing scanpath models.
This might be surprising at first: most scanpath models are implemented for generating complete scanpaths (e.g., for evaluation with scanpath metrics like ScanMatch), not for computing conditional fixation predictions given a previous scanpath history
However, almost all scanpath models assume that there is an internal model state which is build over the previously sampled fixations.
From the model state a priority map is computed from which the next fixation is generated, which, in turn, updates the internal model state (see Algorithm \ref{alg:sampling}).
Computing a conditional fixation prediction in such model simply requires updating the internal state given previous scanpath fixations instead of sampled fixations and then returning the priority map that the model would use to select the next fixation.
Algorithm \ref{alg:conditionalprediction} illustrates that this requires only minimal changes to Algorithm \ref{alg:sampling}.
Models which predict a next fixation location without a priority map, e.g., as the result of solving a differential equation \parencite{zancaGravitationalLawsFocus2019} can be easily converted into priority map models by encoding the predicted fixation location, e.g., with a Gaussian centered at the predicted fixation location.
In this situation, the AUC of the next ground truth fixation location in the Gaussian priority map corresponds to the squared euclidean distance between the predicted fixation location and the next ground truth fixation.

\subsection{Other means of quantifying performance}
\label{sec:scanpath_metrics_other}

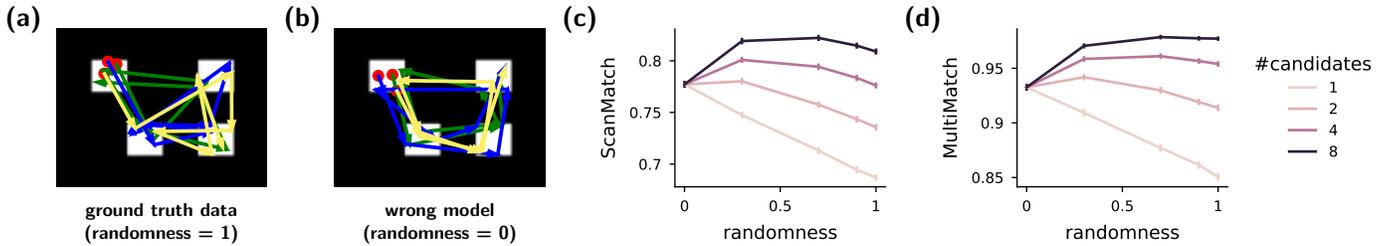
\begin{figure*}
    \begin{center}
        \begin{tikzpicture}
            \coordinate (hsep) at (0.3, 0);
            \coordinate (vsep) at (0, -5);
            \coordinate (labelsep) at (-0.5, -0.0);
            \tikzset{label/.style={font=\sffamily\bfseries}};
            \tikzset{anchor=north west};
            
            \node (examplegt) at (0, 0) {\includegraphics[scale=1.0]{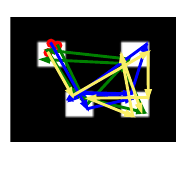}};
            \node (exampleother) at ($ (examplegt.north east) + (hsep) $) {\includegraphics[scale=1.0]{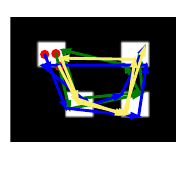}};
            
            \node (scanmatch) at ($ (exampleother.north east) + (hsep) $) {\includegraphics[scale=1]{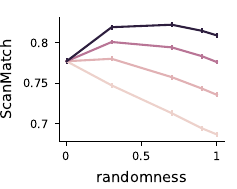}};
            
            \node (multimatch) at ($ (scanmatch.north east) + (hsep) $) {\includegraphics[scale=1]{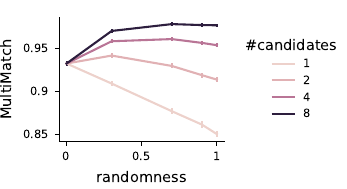}};

            
            
            \node[label] at ($ (examplegt.north west) + (labelsep) $) {(a)};
            \node[label] at ($ (exampleother.north west) + (labelsep) $) {(b)};
            \node[label] at ($ (scanmatch.north west) + (labelsep) $) {(c)};
            \node[label] at ($ (multimatch.north west) + (labelsep) $) {(d)};
            
            \node[label, anchor=north, scale=0.7] (examplegtcaption) at ($ (examplegt.south) + (0.0, 0.8) $) {ground truth data};
            \node[label, anchor=north, scale=0.7] (examplegtcaption2) at ($ (examplegtcaption.south) + (0.0, 0.1) $) {(randomness = 1)};
            \node[label, anchor=north, scale=0.7] (exampleothercaption) at ($ (exampleother.south) + (0.0, 0.8) $) {wrong model};
            \node[label, anchor=north, scale=0.7] (exampleothercaption2) at ($ (exampleothercaption.south) + (0.0, 0.1) $) {(randomness = 0)};
        \end{tikzpicture}
    \end{center}
    \caption{Scanpath similarity metrics can score wrong models better than the generating model. (a) We assume there are four salient objects and that observers usually cycle through them in a fixed order, but sometimes make saccades to a random object.
    (b) We evaluate models which predict the random saccades occur less often or, in the extreme case, never.
    ``Randomness'' indicates how less likely random saccades are compared to the ground truth model: randomness=1 is the ground truth model and randomness=0 is a model which predicts fixations to be strictly cycling through the four objects.
    We show three generated scanpaths (green, blue, yellow) from the ground truth model (a) and the least random model (b).
    (c) ScanMatch scores of the different models ordered by randomness, depending on how many model scanpaths we evaluate per ground truth scanpath (different lines). For every number of candidates there are wrong models (randomness $<$ 1) which perform better than the correct model (randomness$=$1).
    (d) Same as (c), but evaluating MultiMatch instead of ScanMatch. For simplicity, we average over the different MultiMatch components. The results are qualitatively the same for all components. Results are average scores over 10000 scanpaths of 8 fixations each sampled from the ground truth model. Error bars indicate bootstrapped 95\% confidence intervals for the mean}
    \label{fig:problemscanpathsimilarity}
\end{figure*}

Opposed to the evaluation approach we just outlined, in the scanpath literature, the most common approach for evaluating prediction quality is to generate new scanpaths from the model and compare them to actual human scanpaths.
Many different methods and metrics have been proposed for this comparison

Especially in earlier papers, researchers often simply \textit{visually inspected} scanpaths (e.g, \cite{notonScanpathsSaccadicEye1971,brockmannEcologyGazeShifts2000,boccignoneModellingGazeShift2004}).
Additionally, some researchers \textit{compare some statistics} such as saccade length distribution of the sampled scanpaths to those of real scanpaths (e.g., \cite{engbertSpatialStatisticsAttentional2015}).
Some researchers compute average fixation densities from the sampled scanpaths and compare them to empirical fixation densities (e.g., \cite{zancaGravitationalLawsFocus2019, sunStatisticalModelingSaccadic2014,yangPredictingGoalDirectedHuman2020}).

In order to make the evaluation more principled, many researchers have proposed \textit{distance metrics between scanpaths} for computing distances between sampled and ground truth scanpaths.
Typically, these metrics use some means of aligning the fixations in two scanpaths (possibly introducing gaps in either one) and then compute a distance between the aligned fixations.
Here we mention some of the more commonly used such metrics, see \cite{andersonComparisonScanpathComparison2015} for a extensive overview of scanpath distance metrics.
The different metrics differ in if and how they simplify scanpaths, how they align them and when two fixations are considered a close match.
\textcite{brandtSpontaneousEyeMovements1997} suggested to divide the image into a grid and compute string edit distances between scanpaths, considering fixation equal if they fall into the same grid cell.
\textcite{cristinoScanMatchNovelMethod2010} proposed the \textit{ScanMatch} metric which uses the Needleman-Wunsch algorithm to align the scanpaths taking distances between ROIs or fixations as well as fixation duration into account.
The \textit{MultiMatch} \parencite{jarodzkaVectorbasedMultidimensionalScanpath2010} metric first simplifies scanpaths, then aligns them using the Dijkstra algorithm and finally computes multiple distances as differences between saccade vectors, saccade lengths, fixation positions, saccade directions and fixation durations.

While these scanpath metrics are commonly used in the field, they have some problems:
Scanpaths are essentially a series of decisions to saccade to a new fixation location.
Testing whether the model and a human subject, at some point in a scanpath, make a similar fixation (i.e., make a similar decision) only makes sense if this decision is made under the same circumstances.
When comparing the fixations in two scanpaths, this doesn't hold anymore starting from the second fixation.
Aligning the scanpaths can compensate this only to a certain degree.
Evaluating conditional predictions, as we detailed above, compares the decisions of models and human subjects under identical conditions.
Besides this conceptual problem, most common scanpath metrics depend on many hyperparameters that have to be choosen.
A possible scanpath simplification strategy has to be choosen, a grid cell size or ROIs have to be selected, the alignment strategy itself has many tunable parameters such as the gap penalty and the a fixation distance metric.
Also, it is common to sample multiple scanpaths for each ground truth scanpath and choose the most similar one for evaluation.
How many such scanpaths are sampled is another hyperparameter that crucially affects model scores.
All of this means that prediction scores in the literature are very hard to compare. 

But there is a much more severe problem when using scanpath similarity metrics to evaluate scanpath model performance:
incorrect models can score better than the real model which generated the ground truth data.
This is illustrated in Figure \ref{fig:problemscanpathsimilarity}.
Imagine an image with four salient objects, where observers cycle through them in a fixed order, but sometimes make saccades to a random object.(Figure \ref{fig:problemscanpathsimilarity}a).
Let's further assume that besides the model of the ground truth scanpath distribution which we just described, we have other models which predict the random saccades to be less likely or even impossible (Figure \ref{fig:problemscanpathsimilarity}b), with other words: these models are incorrect.
These incorrect models are more deterministic or less random than the ground truth models.
If we evaluate these incorrect models and the ground truth model under ScanMatch or MultiMatch, then we see that incorrect models achieve higher scores  than the real, ground truth model (Figure \ref{fig:problemscanpathsimilarity}c, d).
The effect is most extreme when we sample only one model scanpath per ground truth scanpath and compute the similarity score between these two scanpaths (\#candidates=1 in Figure \ref{fig:problemscanpathsimilarity}c, d). 
The more deterministic the model is, the higher the score, with the least random model scoring 7-8\% better than the ground truth model -- which is actually the correct, perfect model.
When sampling multiple model scanpaths per ground truth scanpath and using only the one with the best similarity score, as often done in the literature (\#candidates$>$1 in Figure \ref{fig:problemscanpathsimilarity}c, d), the effect becomes less pronounced, but there are still always some models that score higher than the ground truth model.
At the same time, simply sampling more scanpaths introduces new problems: when sampling, for example, eight model scanpaths per ground truth scanpath, a model could generate completely unrealistic scanpaths half of the time and would hardly be penalized for it.

The underlying problem here is that in the most extreme case, there might exist one scanpath that is more similar to ground truth scanpaths on average, than one ground truth scanpath is similar to another. By always generating this one scanpath, a model would score higher than by sampling scanpaths from the ground truth distribution.

Of course, metrics often differ in how they quantify differences between model and ground truth, which can lead to \textit{different suboptimal} models performing best in different metrics, which is a common problem in spatial saliency \parencite{kummererSaliencyBenchmarkingMade2018}.
However, the problem that \textit{wrong} models can score better than the \textit{real} model is qualitatively different and shows that scanpath similarity metrics should in general not be trusted as a measure of quality for scanpath models.

Instead, here we advocate for the evaluation of scanpath-history-dependent predictions (conditional predictions).
Conditional predictions cannot score a wrong model higher than the generating model, because the likelihood of a probabilistic model is the entropy of the generating distribution plus the Kullback-Leibler-Divergence between the generating distribution and the model distribution.


\section{Models of Scanpath Prediction}


\begin{table*}
    \begin{center}
        
\begin{tabular}{lccccc}
  \toprule

  \textbf{model} & \textbf{\makecell{image \\ computable}} & \textbf{\makecell{dependency \\ order}} & \textbf{\makecell{allows \\ conditioning}} & \textbf{\makecell{priority \\map}} & \textbf{prob.}\\\hline
  \textbf{Biologically inspired models}\\
  \phantom{foobar}Itti-Koch \parencite{ittiModelSaliencybasedVisual1998} & $\checkmark$ & $\infty$ & $\checkmark$ & $\checkmark$ & $\times$\\
  \phantom{foobar}\textit{Wang \parencite{wangSimulatingHumanSaccadic2011}} & $\checkmark$ & $\infty$ & $\checkmark$ & $\checkmark$ & $\checkmark$\\
  \phantom{foobar}SceneWalk \parencite{engbertSpatialStatisticsAttentional2015} & $\checkmark$ & $\infty$ & $\checkmark$ & $\checkmark$ & $\checkmark$\\
  \phantom{foobar}MASC \parencite{adeliModelSuperiorColliculus2017} & $\checkmark$ & $\infty$ & $\checkmark$ & $\checkmark$ & $\times$\\
  \phantom{foobar}\textit{LATEST \parencite{tatlerLATESTModelSaccadic2017}} & $\checkmark$ & 2 & $\checkmark$ & $\checkmark$ & $\times$\\
  \phantom{foobar}G-Eymol \parencite{zancaGravitationalLawsFocus2019} & $\checkmark$ & 1 & $\checkmark$ & $\times$ & $\times$\\
  \textbf{Statistically inspired models}\\
  \phantom{foobar}\textit{\cite{brockmannEcologyGazeShifts2000}} & $\checkmark$ & 1 & $\checkmark$ & $\checkmark$ & $\checkmark$\\
  \phantom{foobar}CLE \parencite{boccignoneModellingGazeShift2004} & $\checkmark$ & 1 & $\checkmark$ & $\checkmark$ & $\checkmark$\\
  \phantom{foobar}\textit{SGP \parencite{sunStatisticalModelingSaccadic2014}} & $\checkmark$ & N.A. & $\times$ & $\checkmark$ & $\times$\\
  \phantom{foobar}LeMeur16 \parencite{lemeurIntroducingContextdependentSpatiallyvariant2016} & $\checkmark$ & 1 & $\checkmark$ & $\checkmark$ & $\checkmark$\\
  \phantom{foobar}SaccadicFlow \parencite{clarkeSaccadicFlowBaseline2017} & $\checkmark$ & 1 & $\checkmark$ & $\checkmark$ & $\checkmark$\\
  \phantom{foobar}\textit{HMMs \parencite{coutrotScanpathModelingClassification2018}} & $\times$ & 1 & $\checkmark$ & $\checkmark$ & $\checkmark$\\
  \phantom{foobar}IRL \parencite{xiaPredictingHumanSaccadic2019} & $\checkmark$ & $\infty$ & $\checkmark$ & $\checkmark$ & $\times$\\
  \textbf{Cognitively inspired models}\\
  \phantom{foobar}\textit{\textcite{liuSemanticallyBasedHumanScanpath2013}} & $\checkmark$ & 1 & $\checkmark$ & $\checkmark$ & $\checkmark$\\
  \phantom{foobar}ROI-LSTM \parencite{sunVisualScanpathPrediction2019} & $\checkmark$ & $\infty$ & $\checkmark$ & $\checkmark$ & $\checkmark$\\
  \textbf{Engineered models}\\
  \phantom{foobar}SaltiNet \parencite{assensSaltiNetScanPathPrediction2017} & $\checkmark$ & 1 & $\checkmark$ & $\checkmark$ & $\checkmark$\\
  \phantom{foobar}\textit{PathGAN \parencite{assensPathGANVisualScanpath2019}} & $\checkmark$ & $\infty$ & $\times$ & $\times$ & $\times$\\
  \phantom{foobar}DeepGaze III \parencite{kummererDeepGazeIIIUsing2019} & $\checkmark$ & 2 & $\checkmark$ & $\checkmark$ & $\checkmark$\\

\bottomrule
\end{tabular}

    \end{center}
    \caption{Models of human scanpath prediction. We categorize models depending on where they draw their inspiration from and on how they relate to some important properties where models might differ.
    Models displayed in italic could not be included in our performance evaluation for various reasons. ``prob.`` = probabilistic.
    }
    \label{tab:models}
\end{table*}

In the following, we give an overview of the existing literature on predicting human scanpaths.
We classify models into several categories, based on where they take their inspiration from (e.g., biology or statistics).
In Table \ref{tab:mit1003} we summarize all models with respect to some of their most important properties.
Besides the category of their motivation, we show whether they are image-computable (i.e., can be applied to arbitrary images).
and on how many previous fixations in the scanpath the prediction for the next fixation depends (\textit{dependency order}).
Additionally for each model we check the criteria for conditional prediction maps which we listed above in Section \ref{sec:scanpath_prediction}.
Specifically, we ensure that the model architecture allows to condition the prediction for the next fixation on certain previous fixations,
and whether the model computes an explicit priority map to sample from.
Finally we distinguish between probabilistic models that sample from a conditional fixation distribution and classic models that sample from a conditional priority map, e.g., with WTA.

\subsection{Biologically inspired models}

Many models take their inspiration from results in neuroscience and vision science.

\smallparagraph{Itti-Koch:} The seminal model of Itti-Koch \parencite{ittiModelSaliencybasedVisual1998} is nowadays usually used as a spatial saliency model predicting saliency maps.
But the model actually includes a fixation selection mechanism where a priority map is evolved with a biologically inspired winner-takes-all-network \parencite{kochShiftsSelectiveVisual1985} until a winner is selected.
Then the WTA map is modulated with inhibition-of-return and the winner-takes-all-network starts processing again.

\smallparagraph{\citeauthor{wangSimulatingHumanSaccadic2011}:} The model of \textcite{wangSimulatingHumanSaccadic2011} computes in each step response maps of sparse coding filters applied to a foveal image.
These response maps then update response maps in a visual working memory.
From the differences between the response maps in visual working memory and the response maps of the full image, a residual perceptual information map (RPI) is computed, which encodes in which image areas the visual working memory is missing the most information.
The next fixation is selected to be the maximum of the RPI under certain saccade length constraints.

\smallparagraph{SceneWalk:} The \textit{SceneWalk} model \parencite{engbertSpatialStatisticsAttentional2015, schuttLikelihoodbasedParameterEstimation2017} is a mechanistic model that implements an attention and inhibition-of-return mechanism on top of a saliency map.
The attention mechanism reinforces the area around the current fixation and has a short decay time, while the inhibition-of-return mechanism inhibits a smaller area around the current fixation with a longer decay time to predict a probability distribution over possible next fixation locations.
The model was first published with hand-chosen parameters \parencite{engbertSpatialStatisticsAttentional2015}.
Later, \citeauthor{schuttLikelihoodbasedParameterEstimation2017} optimized the parameters for maximum likelihood.

\smallparagraph{MASC:} The \textit{model of attention in the superior colliculus} (MASC, \cite{adeliModelSuperiorColliculus2017}) applies results from neurophysiology to saccade prediction. In order to predict a saccade target, the model first applies a retina transformation to the input image that blurs the image with a blur size that depends locally on the distance to the last fixation.
For this retina-transformed image, subsequently a priority map is computed which is then mapped into superior colliculus space. Here, multiple stages of local averaging (corresponding to different computations known to take place in the brain) are applied before the next fixation location is selected as the location of the maximum.

\smallparagraph{LATEST:} The \textit{Linear Approach to Threshold Explaining Space and Time} (LATEST, \cite{tatlerLATESTModelSaccadic2017}) assumes that for each possible saccade target a decision unit accumulates evidence in favor of saccading there with a rate depending on factors like low-level and semantic content at that location as well as biases like incoming and outgoing saccade direction and saccade amplitude.
When one of the units hits the decision threshold, a saccade to the corresponding location is initiated.
This framework makes predictions both about the location as well as about the timing of saccades.
The authors model fit the parameters of the decision process only to predict the timing of saccades and show that the model as a consequence also predicts fixation locations well.

\smallparagraph{G-Eymol:} The \textit{Gravitational Eye movement laws} (G-Eymol, \cite{zancaGravitationalLawsFocus2019}) models gaze movement by a second order differential equation system that takes into account the gradient of brightness, the gradient of optical flow (which is zero for static images) and an inbition-of-return potential which is updated at regular time intervals.
Scanpaths are generated by simulating the gaze position according to the differential equation and then applying fixation-detection algorithms to extract fixations.

\subsection{Statistically inspired models}

There is another large class of models that don't take their inspiration from biology, but instead try to reproduce certain statistical properties of human scanpaths.

\smallparagraph{Brockmann \& Geisel:} \textcite{brockmannEcologyGazeShifts2000} hypothesized saccades targets to be distributed proportionally to the product of a saliency potential and a function depending on the distance to the last fixation which describes the jump distribution. They showed that this jump distribution is modeled well by a Cauchy distribution.

\smallparagraph{CLE:} The \textit{constrained levy exploration} (CLE) model \parencite{boccignoneModellingGazeShift2004} models saccades with a Levy flight (i.e., using a Cauchy distribution for saccade lengths). However, unlike in a pure levy flight, the CLE model modulates the jump distribution with a saliency map: the starting point of the jump distribution is moved from the last fixation location along the gradient of a potential from the saliency map. Additionally, saccade targets with higher saliency are given a higher probability of being selected.

\smallparagraph{SGP:} The \textit{Super Gaussian Component Pursuit} model \parencite{sunStatisticalModelingSaccadic2014} takes inspiration from component analysis techniques: it computes a super Gaussian component analysis over patches from the image and selects the $k$th fixation to be at the patch with highest response to the $k$th super Gaussian component.

\smallparagraph{LeMeur15:} The model of \textcite{lemeurSaccadicModelEye2015} use a product of a saliency potential and a jump distribution, just like \textcite{brockmannEcologyGazeShifts2000} and \textcite{boccignoneModellingGazeShift2004}. However, unlike these previous models, they don't use a parametric jump distribution but approximate the jump distribution from the data using nonparametric methods.

\smallparagraph{LeMeur16:} The authors of the model LeMeur15 published an extended version a year later \parencite{lemeurIntroducingContextdependentSpatiallyvariant2016}.
The most relevant difference is that the jump distribution is not the same all over the image.
Instead, the image is separated into a grid of 9 areas and for each area the jump distribution is learned separately.

\smallparagraph{SaccadicFlow:} The \textit{SaccadicFlow} model \parencite{clarkeSaccadicFlowBaseline2017} implements a Gaussian jump distribution that depends in a polynomial way on the previous fixation distribution and fits the parameters of this dependency on data. Most notably, the model is independent of the image content and is intended to serve as a baseline model capturing oculumotor biases.

\smallparagraph{HMM:} \textcite{coutrotScanpathModelingClassification2018} model scanpaths with image and task dependent hidden markov models (HMM) where each state is a 2d Gaussian in the image with certain transition probabilities between states.
The are fitted to each image and task individually.

\smallparagraph{IRL:} The \textit{iterative representation learning} (IRL; \cite{xiaPredictingHumanSaccadic2019}) model defines the salience of an image location to be the residual of an autoencoder that has been learned using patches around previously fixated locations.
This means that the model is more likely to look at image locations that don't adhere to the statistics of previously visited image areas.
The autoencoder is initialized using random patches from the image.
After each fixation the training set for the autoencoder is augmented with patches around the last fixated location and the autoencoder is finetuned on this new dataset.
The model combines the computed saliency with a center bias and an inhibition of return mechanism to compute a priority map from which the location of the maximum is picked for the next fixation.

\subsection{Cognitively inspired models}

While for a long time it was assumed that at least free-viewing eye movements are mainly driven by low-level features such as edges and contrast, by now it is accepted that high-level content such as objects and faces strongly attract attention \parencite{einhauserObjectsPredictFixations2008,borjiObjectsNotPredict2013,einhauserObjectsSaliencyReply2013,kummererUnderstandingLowHighLevel2017}.
This raises the question which cognitive effects affect gaze placement.
Researchers have proposed several models that incorporate such effects.

\smallparagraph{\citeauthor{liuSemanticallyBasedHumanScanpath2013}:} The model by \textcite{liuSemanticallyBasedHumanScanpath2013} models fixation selection as a product of low-level saliency effects, semantic effects and spatial effects.
The effect of low-level saliency is modeled as a probability depending on the norm of the difference between low-level features at the last fixation location and a candidate target location.
For semantic effects, the authors learn a a hidden Markov model whose observations are representations of image regions in terms of a Bag-of-Visual-Words model using SIFT features.
Finally, spatial effects are modeled using a Levy flight distribution.

\smallparagraph{IOR-ROI-LSTM:} The IOR-ROI-LSTM model \parencite{sunVisualScanpathPrediction2019} uses deep learning models to encode an image into deep features and predicted semantic masks.
This data is then feed into a recurrent neural network architecture which uses channel-wise attention to inhibit certain image features and predict the next region of interest via a mixture of Gaussians.
This prediction is combined with oculomotor biases to select the next fixation location.

\subsection{Engineered models}

Finally, there is a class of models which we call ``engineered models''.
Instead of implementing mechanisms taken from biological or statistical ideas, they are deep learning based models that are fitted to the data.

\smallparagraph{SaltiNet:} \textcite{assensSaltiNetScanPathPrediction2017} use a deep neural network to predict a spatiotemporal saliency volume which is then combined with a jump distribution prior to predict saccades.

\smallparagraph{PathGAN:} \textcite{assensPathGANVisualScanpath2019} proposed a generative adversarial network architecture \parencite{goodfellowGenerativeAdversarialNets2014} where the generator is a recurrent neural network predicting scanpaths as sequences of spatiotemporal coordinates and the discriminator tries to discriminate these generated scanpaths from ground truth human scanpaths.

\smallparagraph{DeepGaze III:} DeepGaze II \parencite{kummererUnderstandingLowHighLevel2017} was the best performing spatial saliency model on the MIT Saliency Benchmark for several years.
    Later, the authors proposed an extension towards scanpath prediction \parencite{kummererDeepGazeIIIUsing2019}.
DeepGaze III uses deep features from the VGG19 network in a readout network of $1 \times 1$ convolutions to compute a spatial saliency map for an input image and encodes the previous two fixation locations in spatial feature maps.
Saliency map and scanpath features are then combined via multiple additional layers of $1 \times 1$ convolutions to yield a distribution predicting the location of the next fixation in the scanpath.

\section{Methods}

\subsection{Datasets}

We use several datasets to conduct our experiments on.
The \textbf{MIT1003} dataset \parencite{juddLearningPredictWhere2009} consists of 1003 images of mainly color natural scenes with a longer side of 1024 pixels.
The authors of the dataset collected eye movements from 15 subjects with 3 seconds presentation time and make scanpaths of fixations available.
Unlike most other works, we do not exclude the initial central fixation from the scanpaths since we want to allow models to model the influence of the initial fixation on later fixations.
In the few cases where the initial fixation is marked as invalid in the eye tracking data, we replace it with a central fixation.
We also evaluate the \textit{MIT300} dataset \parencite{juddBenchmarkComputationalModels2012}, which is the holdout dataset for the MIT1003 dataset.
It consists of data from 45 subjects on 300 images under the same experimental conditions as in MIT1003.
Since it's the benchmark dataset for the MIT/Tuebingen Saliency Benchmark, the fixations are not public to avoid overfitting.
We apply the same preprocessing as for the MIT1003 dataset

As a second dataset, we use the \textbf{CAT2000} dataset \parencite{borjiCAT2000LargeScale2015}.
It's public training part consists of 2000 images of different aspect ratios, all padded to have a size of 1920x1200 pixels.
The images are from 20 different categories, such as indoor, outdoor natural, outdoor manmade, satellite and fractal.
The authors collected fixations from 18 subjects with 5 seconds presentation time.
Since the initial forced fixation is not contained in the provided scanpaths, we add a initial central fixation to each scanpath.
We found several instances of apparently repeated fixations (the exact same fixation locations consecutively in a scanpath).
In these cases we removed the duplicate fixations from the scanpaths.
In addition to the public training set of CAT2000, we also evaluate the nonpublic test set, which consists of fixation data of 24 subjects for another 2000 images from the same categories.

In all datasets, each trial of a subject begins with a initial forced central fixation.
We include this initial forced fixation in the scanpath history that the models are conditioned on, but we don't evaluate how well models predict it, since it is not voluntary.

\begin{table*}[t]
    \begin{center}
        \begin{tabular}{lrrrrcrrrr}
            \toprule
            & \multicolumn{4}{c}{\textbf{MIT1003}} & \quad & \multicolumn{4}{c}{\textbf{MIT300}}\\
            \textbf{Model} & \textbf{LL} & \textbf{IG} & \textbf{AUC} & \textbf{NSS}
            && \textbf{LL} & \textbf{IG} & \textbf{AUC} & \textbf{NSS} \\\hline
            \begin{filecontents*}{data/table_MIT1003_MIT300/table.csv}
Model|TLL|TIG|TAUC|TNSS|VLL|VIG|VAUC|VNSS
Itti\&Koch (with WTA network)|||0.473|0.271|||0.474|0.253
uniform|-0.000|-0.906|0.500|0.000|-0.000|-0.782|0.500|0.000
Itti\&Koch (purely spatial)|||0.545|0.480|||0.540|0.408
MASC (Itti\&Koch)|||0.635|0.545|||0.636|0.537
G-Eymol|||0.662|0.553|||0.654|0.525
STAR-FC|||0.662|0.581|||0.659|0.576
MASC (DG2)|||0.713|1.022|||0.704|0.962
CLE (Itti\&Koch)|0.217|-0.689|0.720|0.980|0.128|-0.655|0.712|0.993
IOR-ROI-LSTM|-46.821|-47.727|0.744|0.457|-49.212|-49.995|0.730|0.420
IRL|||0.748|1.113|||0.747|1.091
SaltiNet|0.720|-0.186|0.790|1.138|0.687|-0.095|0.780|1.079
LeMeur16 (GBVS)|0.038|-0.868|0.782|1.110|0.100|-0.682|0.783|1.077
center bias|0.906|0.000|0.801|1.263|0.782|0.000|0.783|1.096
fixation number dependent centerbias|0.950|0.044|0.801|1.278|0.838|0.056|0.784|1.117
CLE (constant saliency)|0.872|-0.034|0.823|1.031|0.821|0.039|0.815|1.070
Saccadic Flow|1.170|0.264|0.843|1.603|1.070|0.287|0.835|1.491
LeMeur16 (DG2)|0.596|-0.310|0.853|2.270|0.581|-0.201|0.850|2.123
DeepGaze II|1.870|0.965|0.884|2.528|1.724|0.942|0.873|2.337
SceneWalk (DG2)|1.968|1.062|0.888|2.705|1.839|1.056|0.881|2.547
CLE (DG2)|1.765|0.860|0.897|1.488|1.640|0.857|0.888|1.531
CLE (DG2, optimized parameters)|1.801|0.896|0.897|1.448|1.686|0.904|0.888|1.490
SceneWalk (DG2, finetuned)|2.048|1.142|0.898|2.695|1.923|1.141|0.890|2.530
Spatial Gold Standard (subject LOO)|2.120|1.215|0.901|2.853|2.117|1.334|0.899|2.859
CCN-DeepGaze III|2.257|1.351|0.907|2.992|2.171|1.388|0.902|2.886
Spatial Gold Standard (joint)|2.762|1.857|0.949|3.393|2.555|1.773|0.936|3.181
\end{filecontents*}
\csvreader[late after line=\\,head to column names,separator=pipe]{data/table_MIT1003_MIT300/table.csv}{}%
            {\Model & \TLL & \TIG & \TAUC & \TNSS && \VLL & \VIG & \VAUC & \VNSS }%
            \bottomrule
        \end{tabular}
    \end{center}
    \caption{Model performances on the public MIT1003 and the heldout MIT300 dataset. If applicable, for each model we specify which model has been used as internal saliency map model.}
    \label{tab:mit1003}
\end{table*}

\begin{table*}[t]
    \begin{center}
        \begin{tabular}{lrrrrcrrrr}
            \toprule
            & \multicolumn{4}{c}{\textbf{CAT2000 train}} & \quad & \multicolumn{4}{c}{\textbf{CAT2000 test}}\\
            \textbf{Model} & \textbf{LL} & \textbf{IG} & \textbf{AUC} & \textbf{NSS}
            && \textbf{LL} & \textbf{IG} & \textbf{AUC} & \textbf{NSS} \\\hline
            \begin{filecontents*}{data/table_CAT2000/table.csv}
Model|TLL|TIG|TAUC|TNSS|VLL|VIG|VAUC|VNSS
Itti\&Koch (with WTA network)|||0.379|-0.003|||0.380|-0.006
uniform|0.000|-1.439|0.500|0.000|0.000|-1.391|0.500|-1.000
Itti\&Koch (purely spatial)|||0.526|0.267|||0.525|0.255
MASC (Itti\&Koch)|||0.556|0.248|||0.557|0.252
STAR-FC|||0.610|0.303|||0.613|0.319
IRL|||0.612|0.545|||0.615|0.557
G-Eymol|||0.618|0.227|||0.618|0.225
MASC (DG2)|||0.676|0.835|||0.676|0.829
IOR-ROI-LSTM|-53.386|-54.825|0.687|0.222|-53.128|-54.519|0.687|0.228
CLE (Itti\&Koch)|1.365|-0.074|0.818|2.697|1.324|-0.067|0.815|2.603
LeMeur16 (GBVS)|-0.230|-1.668|0.820|1.362|-0.275|-1.666|0.818|1.349
SaltiNet|0.918|-0.521|0.823|1.329|0.913|-0.478|0.822|1.323
LeMeur16 (DG2)|-0.700|-2.139|0.844|1.711|-0.731|-2.122|0.842|1.682
center bias|1.439|0.000|0.849|2.156|1.391|0.000|0.844|2.087
fixation number dependent centerbias|1.570|0.132|0.850|2.350|1.524|0.133|0.846|2.277
SceneWalk (DG2)|1.806|0.367|0.853|2.708|1.791|0.400|0.853|2.673
DeepGaze II|1.501|0.062|0.866|2.001|1.475|0.084|0.864|1.962
Spatial Gold Standard (subject LOO)|1.882|0.443|0.885|2.517|1.864|0.473|0.884|2.488
CLE (constant saliency)|1.769|0.330|0.897|3.290|1.736|0.345|0.895|3.162
Saccadic Flow|1.721|0.282|0.907|2.199|1.701|0.310|0.905|2.173
CCN-DeepGaze III|2.291|0.852|0.913|2.924|2.255|0.864|0.911|2.874
CLE (DG2)|2.581|1.142|0.915|3.453|2.531|1.140|0.912|3.327
Spatial Gold Standard (joint)|2.299|0.860|0.923|2.868|2.194|0.803|0.916|2.743
\end{filecontents*}
\csvreader[late after line=\\,head to column names,separator=pipe]{data/table_CAT2000/table.csv}{}%
            {\Model & \TLL & \TIG & \TAUC & \TNSS && \VLL & \VIG & \VAUC & \VNSS }%
            \bottomrule
        \end{tabular}
    \end{center}
    \caption{Model performances on the CAT2000 dataset both for the public training and the heldout test set. If applicable, for each model we specify which model has been used as internal saliency map model.}
    \label{tab:cat2000}
\end{table*}

\subsection{Models}

\subsubsection{Baseline models}

We included a number of baseline models in our evaluation to put the numbers of other models into perspective

\smallparagraph{Uniform:} The uniform model predicts fixations to be uniformly and independently distributed over the image.

\smallparagraph{Center bias:} It is known that people have a tendency to look at the center of images \parencite{tatlerCentralFixationBias2007}, the so called \textit{center bias}.
The center bias model uses for each image Gaussian kernel density estimate over all fixations on other images in the same dataset.
It quantifies how well fixations can be predicted without knowing the presented image and the previous scanpath history.

\smallparagraph{Fixation number dependent center bias} The center bias is known to decay over time \parencite{tatlerCentralFixationBias2007}.
The fixation number dependent center bias model estimates the center bias separately depending on the number of the fixation in the scanpath.
For the MIT1003 and MIT300 datasets, the scanpaths were split into intervals as in 1, 2, 3--5, 6--$\infty$.
For the CAT2000 dataset, the scanpaths were split into intervals as in 1, 2, 3, 4, 5--6, 7--10, 11--15, 16--$\infty$.

\smallparagraph{Spatial gold standard:} The spatial gold standard model predicts fixations of each subject by means of a Gaussian KDE over all fixations of other subjects on the same image \parencite{kummererInformationtheoreticModelComparison2015}.
More precisely, our gold standard model is a mixture of a Gaussian KDE over all fixations of other subjects on the same image, a uniform component and a component given by the center bias (see above).
This model encodes the assumption, that fixations will usually be close to fixations from other subjects, but sometimes just randomly placed in the center of the image or uniformly randomly placed.
The mixture weights as well as the bandwidth of the Gaussian KDE have been selected to yield maximum likelihood on the dataset in a leave-one-subject-out crossvalidation paradigm.
We report both leave-one-subject-out performance as well as joint performance

\subsubsection{Published Models}
\label{sec:method_published_models}

We evaluated a range of existing scanpath models.
Basend on our evaluation protocol, criteria for including a model were that (a) we could get the original implementation online or from the authors, that (b) the model explicitly included previous fixation locations in it's internal state, and that (c) the model is image-computable
This excluded several models, e.g., PathGAN which recurrently evolves internal LSTMs that cannot be fed with actual fixation coordinates instead of sampled ones (b) or Coutrot's HMM model \parencite{coutrotScanpathModelingClassification2018} which is not image-computable (c) but has to be trained for each image on eye movement data.

For our evaluation, we need models to compute conditional priority maps given a previous scanpath history.
For models which don't already do this but allow conditioning, we adapt the implementation to update the internal model state from given previous fixations instead of sampled ones as discussed in Section \ref{sec:modifying_models}.
If a model doesn't predict a priority map but only a fixation position, we encode the predicted fixation location into a priority map with a Gaussian centered at the predicted fixation location.
We choose the size of the Gaussian to yield high NSS performance (AUC is invariant under the size of the Gaussian).

Many models require an internal spatial saliency map model.
For better comparison, we evaluate such models once with the internal model that the researchers used originally (e.g., GBVS) and once with the state-of-the-art spatial saliency model DeepGaze II.

See Section \ref{sec:appendix_methods_models} in the appendix for details about how we ran and evaluated each model.


\subsection{Model Evaluation}

We evaluate all models under several established metrics that have been commonly used to assess the performance of models of fixation prediction in the past and that are described below.
For computing dataset-level performances, we first average the performances of each fixation per image and then over all images.


\smallparagraph{LL, IG:} The \textit{information gain} (IG, \cite{kummererInformationtheoreticModelComparison2015}) computes the difference in average log-likelihood between a model and a baseline model.
We report information gain relative to a uniform baseline model as LL and information gain relative to the center bias model as IG.
We use logarithms with the base 2, which means that the values are in the unit of bit per fixation.
The IG metric can only be applied to models which predict a conditional fixation distribution (as opposed to just a conditional priority map without any probabilistic meaning).


\smallparagraph{AUC:} The \textit{Area under the Curve} (AUC, \cite{wilmingMeasuresLimitsModels2011}) evaluates the performance of a priority map when used as binary classifier score for classifying locations into fixated and non-fixated.
In the case of conditional priority maps, there is only one true fixation location and the AUC metric essentially measures the rank of the fixated pixel, with an AUC of 100\% corresponding to a perfect prediction and an AUC of 50\% corresponding to chance level.
There are different versions of AUC that mainly differ in how non-fixation locations are choosen.
We use a uniform nonfixation distribution, i.e., we use all pixels as nonfixations.
Another commonly used version of AUC is \textit{shuffled AUC} (sAUC, \cite{tatlerVisualCorrelatesFixation2005}) which uses fixations from other images as nonfixations.
We don't include it here since sAUC penalizes models for modeling the center bias. However, the center bias is partially result of the initial central fixation and saccadic biases which should be captured by scanpath models.

\smallparagraph{NSS:} The \textit{Normalized Scanpath Salience} (NSS, \cite{petersComponentsBottomupGaze2005}) computes the average saliency of fixated locations after normalizing the saliency map to have zero mean and unit variance.

\smallparagraph{Excluded metrics:} There are several other metrics commonly used for evaluating spatial fixation prediction that we don't use here.
The \textit{shuffled AUC} (sAUC, \cite{tatlerVisualCorrelatesFixation2005}) expects model to not include any central fixation bias and penalizes models who exhibit such a bias.
However, the initial central fixation together with finite presentation time and a bias towards short saccades explains part of the central fixation bias and therefore, sAUC would penalize models that correctly model the saccade length distribution.
Therefore we don't evaluate sAUC.
Distribution-based metrics like \textit{Correlation Coefficient} (CC, \cite{ouerhaniEmpiricalValidationSaliencybased2003}), \textit{Similarity} (SIM, \cite{juddBenchmarkComputationalModels2012}) and \textit{Kullback-Leibler Divergence} (KLDiv, \cite{rajashekarPointofgazeAnalysisReveals2004}) need a spatial estimate of the ground truth distribution.
In classic saliency prediction there are many fixations per image that allow to estimate $p(x, y \mid I)$, however, in the scanpath setting we need to estimate $p(x_i, y_i \mid x_0, y_0, \dots, x_{i-1}, y_{i-1}, I)$ for which we usually will have only one single fixation.
Therefore we don't evaluate CC, SIM or KLDiv.



\section{Results}

\subsection{Benchmarking}

\paragraph{MIT1003 and MIT300:} In Table \ref{tab:mit1003} we show the evaluation results for all evaluated models both on the MIT1003 and MIT300 dataset.
We find that DeepGaze III is the best of the evaluated image-computable models with a average log-likelihood of 2.26bit/fix and an AUC of 90.7\%.
Interestingly, DeepGaze III, SceneWalk and CLE are the only models that perform better than the purely spatial DeepGaze II model.
On the other hand, LeMeur16 and MASC surprisingly decrease performance compared to the internally used DeepGaze II model.
For models that have been evaluated with different internal saliency models we see substantial performance differences (e.g., LeMeur16 with GBVS or with DeepGaze II), underlining how important it is to understand which image features attract attention independent of scanpath history.
At the same time, we also see substantial performance differences between models that use the same internal saliency model but differ in how they use it (e.g., MASC with DeepGaze II and SceneWalk with DeepGaze II), demonstrating the importance of correctly modelling the dependencies between fixations.
Notably, there are even cases where model performance drops compared to the purely static saliency model when adding a scanpath dependent fixation selection mechanism.
Finally, the version of the spatial gold standard where we include all fixations in the Gaussian kernel density estimate (as opposed to leave-one-subject-out prediction) is better than all other models although it cannot take scanpath history into account.

\input{data/table_MIT1003_MIT300/correlation_data.tex}
We find that AUC performance and log likelihoods are not perfectly correlated but quite well rank-correlated (Pearson's $\rho$=\correlationdata[AUC-LL][correlation], Spearman's r=\correlationdata[AUC-LL][spearmanr] on MIT300).
Outliers are mainly models that are probabilistic by definition, but that were not trained as fully probabilistic models, e.g. LeMeur16, or IOR-ROI-LSTM.
The case of IOR-ROI-LSTM is especially extreme:
IOR-ROI-LSTM was trained for maximum likelihood, but without the best-of-k sampling step (see Appendix, Section \ref{sec:best-of-k}) used for enforcing oculomotor biases.
This step strongly changes the model distribution towards low probability values which are heavily penalized by log-likelihood but ignored by the only rank-order sensitive AUC metric.
AUC and NSS are better correlated in value and rank (Pearson's $\rho$=\correlationdata[AUC-NSS][correlation], Spearman's r=\correlationdata[AUC-NSS][spearmanr] on MIT300) but there are still some discrepancies which are most likely due to similar issues as above, which are not penalized by NSS as much as by log-likelihood.

\paragraph{CAT2000} The results on CAT2000 overall resemble the results on MIT1003 and MIT300 with some notable exceptions.
Most notably, the baselines are quite different.
The centerbias reaches substantially higher scores, indicating a stronger centerbias in this dataset.
Most likely, this is due to the larger visual appearance of the images in the experimental setup and the fact that many images are padded.
On the other hand, the spatial gold standard is substantially lower, indicating higher variance across subjects.
DeepGaze II as spatial baseline explains only about 25\% of the explainable information gain between centerbias and gold standard.
This might be due to the images in CAT2000 having much higher diversity than the natural scences in MIT1003 where DeepGaze II was trained.
As a consequence, other models don't profit as much from using DeepGaze II as internal saliency model as it is the case for the MIT1003 dataset.
Interestingly, both CLE without saliency (i.e., a simple Levy flight) and Saccadic Flow are better than all models except for DeepGaze III and CLE with DG2.
This indicates that the saccade statistics on CAT2000 are quite different from MIT1003 and these simple models account better for the changed saccade statistics than more sophisticated models that have been fitted on different datasets.
DeepGaze III is better than CLE without saliency and Saccadic Flow, but is still outperformed by CLE with DeepGaze II, showing that DeepGaze III suffers the same problem.


\begin{figure*}
    \newlength\Colsep
    \setlength\Colsep{10pt}
    \noindent\begin{minipage}{\textwidth}
        \begin{minipage}[c][17cm][c]{\dimexpr0.5\textwidth-0.5\Colsep\relax}
            \begin{center}
                \textsf{\textbf{(a) Cases where model predictions differ most\\
                (all fixations)}}\\
                \includegraphics[height=0.65\textheight]{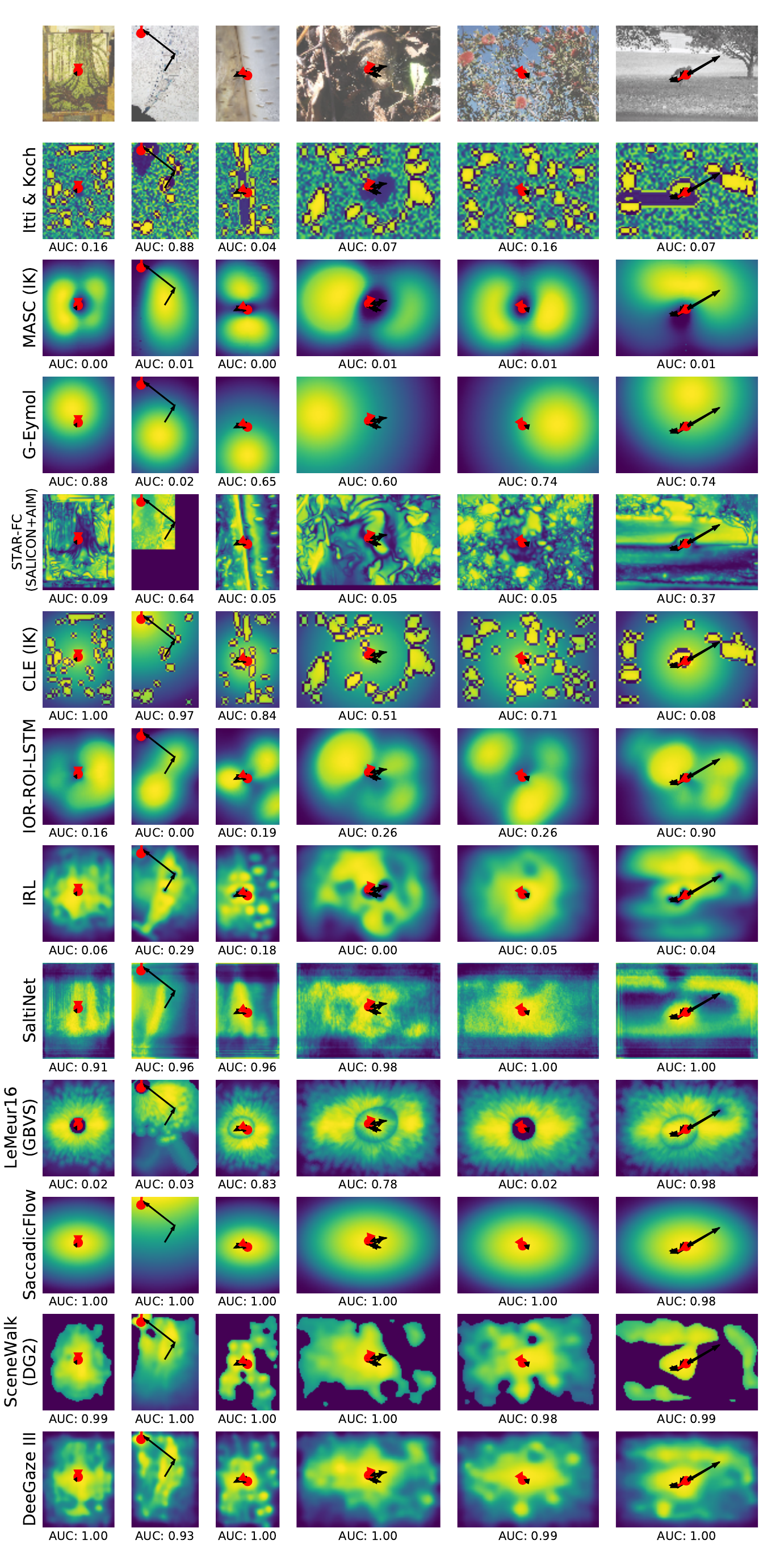}
            \end{center}
        \end{minipage}\hfill
        \begin{minipage}[c][17cm][c]{\dimexpr0.5\textwidth-0.5\Colsep\relax}
            \begin{center}
                \textsf{\textbf{(b) Cases where model predictions differ most\\
                        (excluding short saccades)}}\\
                \includegraphics[height=0.65\textheight]{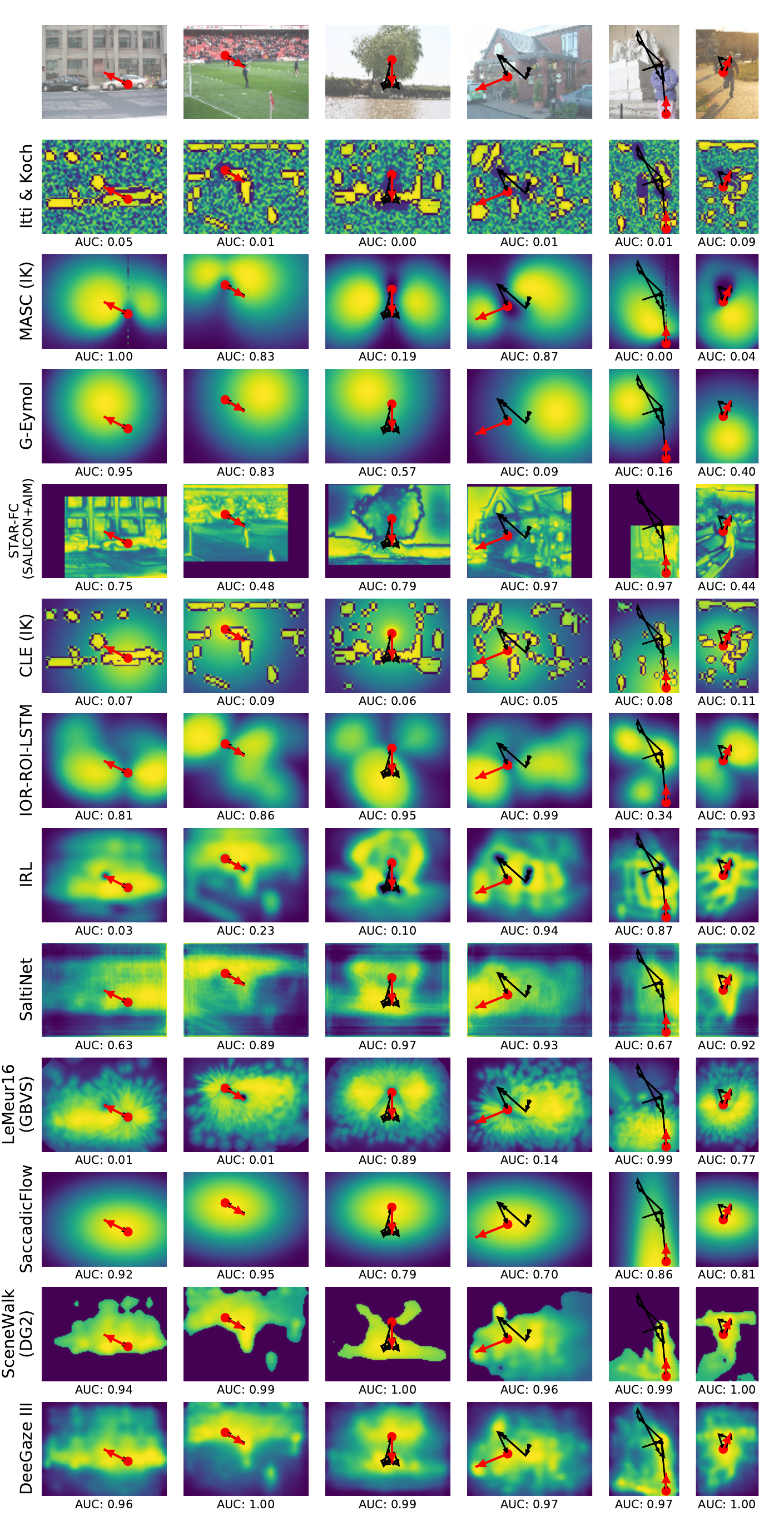}
            \end{center}
        \end{minipage}%
    \end{minipage}

    \caption{Case studies: we show situations where model predictions differ most in terms of AUC performance. Top row: viewed image with previous scanpath (black arrow) and the next saccade (red arrow) which has to be predicted. Other rows: conditional priority maps computed by different models given the previous scanpath history of the fixation.}
    \label{fig:case_studies_most_diverse}
\end{figure*}

\begin{figure*}
    \setlength\Colsep{10pt}
    \noindent\begin{minipage}{\textwidth}
        \begin{minipage}[c][17cm][t]{\dimexpr0.5\textwidth-0.5\Colsep\relax}
            \begin{center}
                \textsf{\textbf{(a) Cases where model predictions differ most\\
                        (excluding saccades to already visited areas)}}\\
                \includegraphics[width=0.82\linewidth]{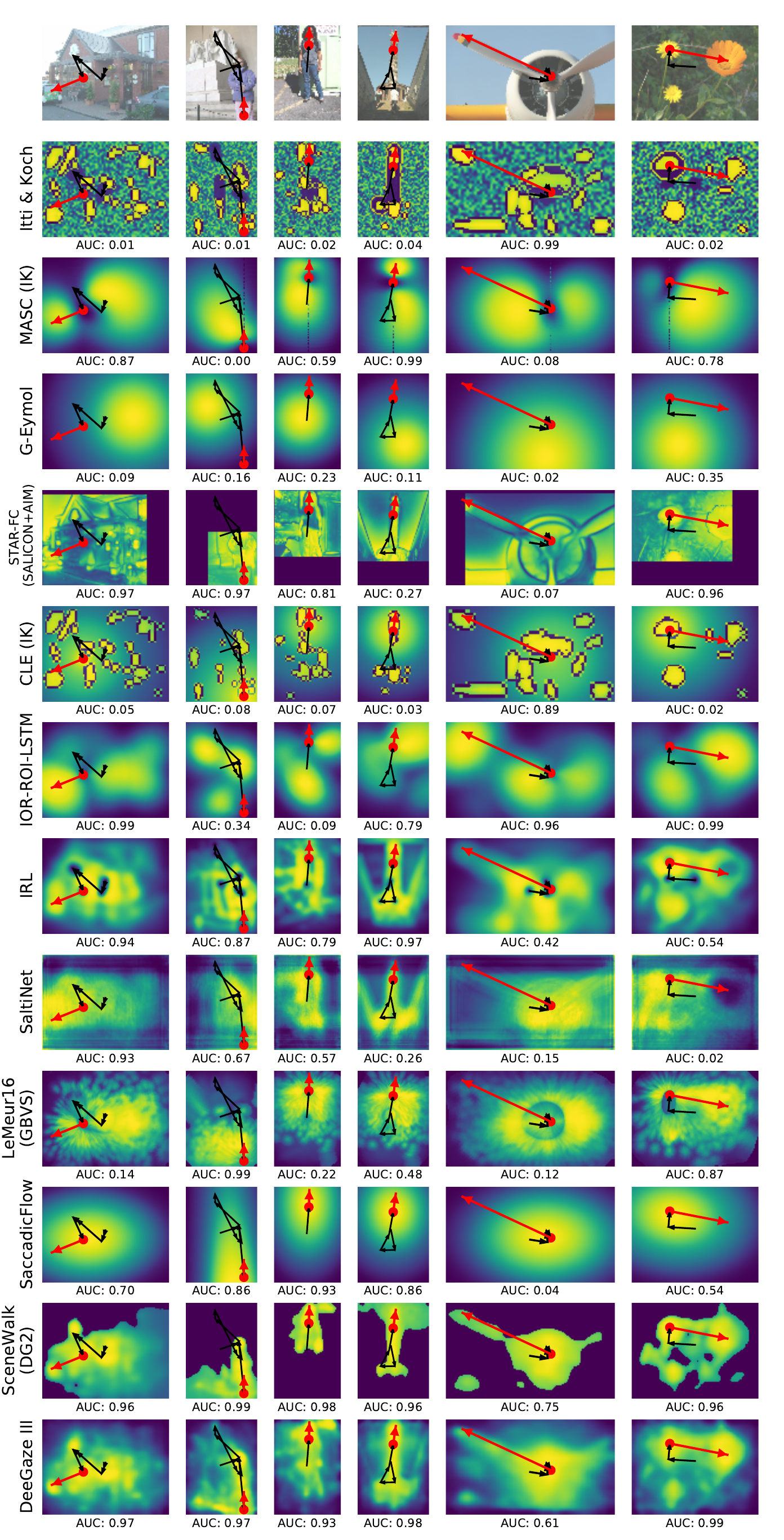}
            \end{center}
        \end{minipage}\hfill
        \begin{minipage}[c][17cm][t]{\dimexpr0.5\textwidth-0.5\Colsep\relax}
            \begin{center}
                \textsf{\textbf{(b) Cases where model predictions differ most\\
                        (excluding saccades to already visited areas)}}\\
                \includegraphics[width=0.82\linewidth]{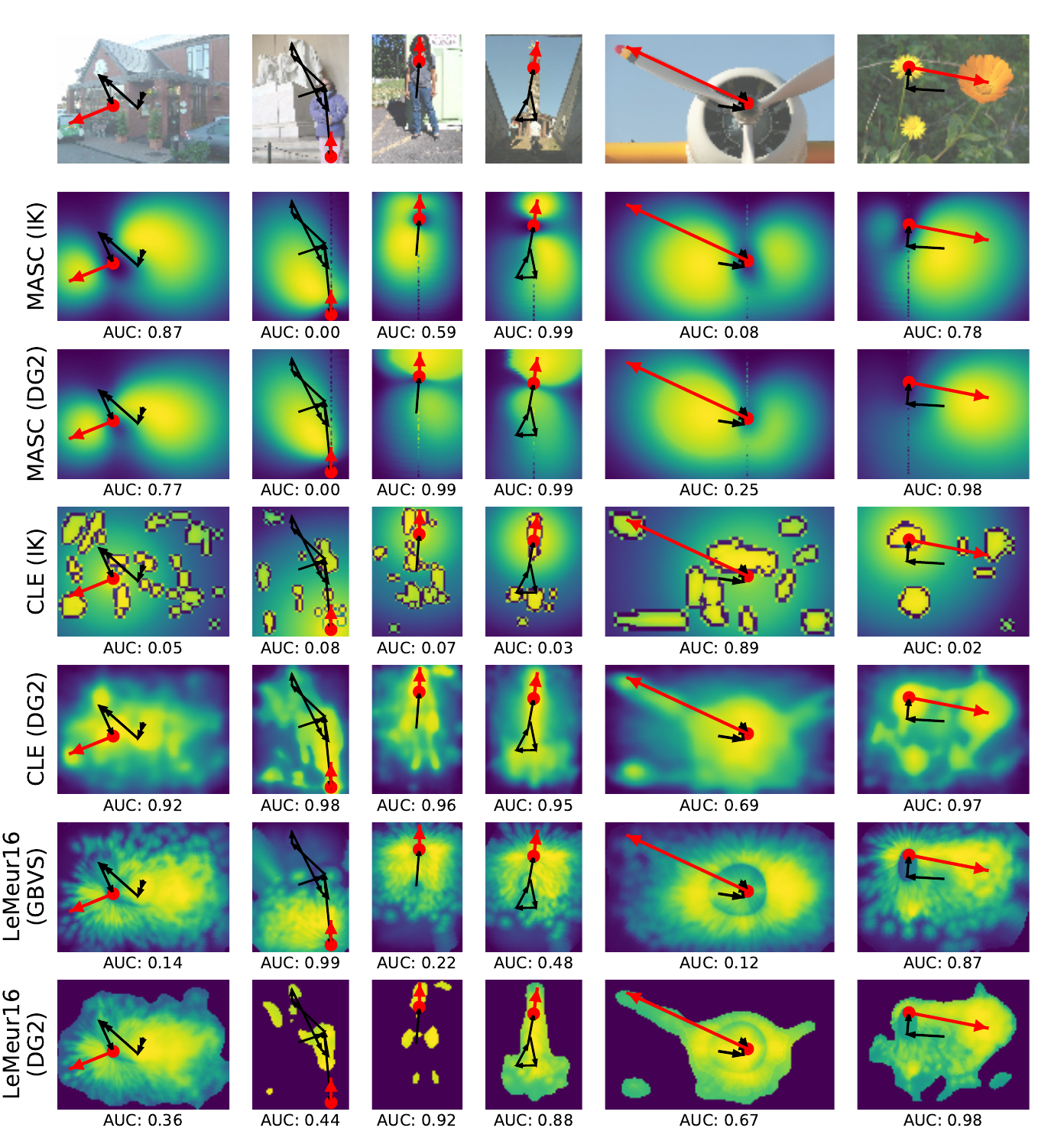}
            \end{center}
        \end{minipage}%
    \end{minipage}
    
    \caption{Case studies: we show situations where model predictions differ most in terms of AUC performance. Top row: viewed image with previous scanpath (black arrow) and the next saccade (red arrow) which has to be predicted. Other rows: conditional priority maps computed by different models given the previous scanpath history of the fixation.}
    \label{fig:case_studies_no_return_most_diverse}
\end{figure*}

\subsection{Case studies}
\label{sec:results_case_studies}

One particular advantage of having models compute conditional predictions for each fixation given the image and the previous scanpath history is that we can analyze the dataset on a per-fixation level and find interesting situations.
For example, we can look for cases where model predictions are especially good, bad or diverse.
In order to quantify how diverse model predictions are, we compute for each fixation the standard deviation of the AUC scores of all models for this single fixation.
In Figure \ref{fig:case_studies_most_diverse}a we show cases where model performances differ most.
For each case (columns), we show the conditional priority maps for many models (rows).
Since AUC is only sensitive to changes in the rank order of values, we normalize all priority maps to have uniform histogram.
We see that the cases where AUC scores differ most all are very short saccades (usually less than two degrees of visual angle).
Models that put a too large inhibition of return suppression on the current fixation location cannot predict those saccades well (e.g., LeMeur16), while models without (or with moderate) inhibition of return mechanisms have a change to predict them well (e.g., SceneWalk).

To exclude those cases from our next analysis, we only look at situations, where the next fixation is more than five degrees of visual angle from the current fixation.
Out of those cases, we show in Figure \ref{fig:case_studies_most_diverse}b again the ones where model performances in AUC differ most.
We find that this time, most cases are return saccades, i.e., saccades back to a location that was already inspected in an earlier fixation of the scanpath.
Models which put a strong inhibition of return suppression on previous fixations score low in theses case (e.g., IRL)

Since in all cases analysed so far, inhibition of return seemed to be the mechanism with most effect on model performance, we finally constrain our analysis to situations where the next fixation is at least five degrees of visual angle from all previous fixations (Figure \ref{fig:case_studies_no_return_most_diverse}a).
This time, it appears all saccades go towards either things that stand out in terms of low-level saliency areas or to high-level content such as objects, text and faces.
Models with a access to high-level information about the scene (e.g, DeepGaze III) can predict the latter case very well while models with only access to low-level features and statistics (e.g., by using GBVS) don't assign high priority to the saccade targets.
By replacing low-level internal saliency map models with the high-level DeepGaze II model, predictions can often improve substantially in these cases (Figure \ref{fig:case_studies_no_return_most_diverse}b)



\section{Discussion}


We presented an extensive summary of the field of scanpath prediction together with a principled evaluation of many models.
There are some scanpath models that reach very high performance.
With DeepGaze III and SceneWalk, these high-performing models include both purely data-driven models and mechanistic models that are inspired from biological results.
The difference between the best performing interpretable model SceneWalk and the best performing deep learning model DeepGaze III is about 0.21 bit/fix compared to 0.18 bit/fix from the purely spatial DeepGaze II model to the SceneWalk:
so far, interpretable models explain about about half of the explainable information in the scanpath history dependency of fixations.
This suggests that there is substantial potential to improve interpretable models with additional mechanisms that account for the gap in performance.
Interestingly, there are also several models that perform worse than underlying spatial saliency models although their mechanisms are informed by biology.
Our fixation-based evaluation of conditional priority map allows us to look into individual failure cases and analyse why some models perform suboptimal.
We find that two major potential sources of performance degeneration are overly strong inhibition-of-return mechanisms (including a penalty for short saccades) and failure to account for effect of high-level content.
While inhibition-of-return is a well researched effect, it's effect on the spatial location of fixations (as opposed to timing effects, see \cite{kleinInhibitionReturn2000}) seems not to be very strong and there are actually many cases where humans make very short saccades or return exactly to previous fixation locations (see also \cite{smithFacilitationReturnScene2009}).

While we consider log likelihoods the most principled and intuitive metric \parencite{kummererInformationtheoreticModelComparison2015}, for the benchmarking purposes that we are interested in here, the AUC metric might be more suitable for multiple reasons:
Many evaluated models are not probabilistic and therefore don't even allow the computation of log likelihoods.
Some other models are probabilistic by design, but authors did not use the full model distribution for fitting, which can affect log-likelihood scores strongly (e.g., IOR-ROI-LSTM, LeMeur16).
AUC is the most robust metric since it only takes the ranking of the values of the conditional priority maps into account.
This means that the metric only sees the iso-contour lines of the conditional model predictions, making it in some sense the most qualitative metric.
This is especially relevant when models are fitted on another dataset.
They might model the qualitative shape of the conditional distributions very well, but might not capture the exact density values for the new dataset perfectly.
In this case, log-likelihood performance will suffer, but AUC performance might be only minimally affected.
The final metric we evaluate, NSS, can be applied to non-probabilistic models as well, but can be very noisy.
It is effectively a z-score, however, the values of the conditional model predictions are usually extremely skewed towards zero.
Therefore, small changes to the conditional model predictions can strongly affect the NSS values.
Overall, we recommend that researchers train and evaluate their models using log-likelihoods and use AUC when comparing to other models that might not have been fitted as probabilistic models or that where fitted on other datasets.

In our considerations, we mostly ignored fixation durations, since only very few models take them into account (e.g., SceneWalk and IOR-ROI-LSTM).
Fixation durations can be relevant in two different ways, for conditioning and for predicting.
First, the durations of past fixations influence future fixations.
This is already possible in our framework by passing not only past fixation locations but also past fixation durations to the model when computing conditional fixation predictions and we already do this for SceneWalk and IOR-ROI-LSTM.
Second, models might predict not only the location of future fixations but also their durations.
This is a straight forward extension to our framework in the case of probabilistic models and log-likelihood evaluation, where we can simple compute the conditional log-likelihood $\log p(x_i, y_i, d_i \mid x_0, y_0, d_0, \dots, I)$, where $d_k$ are fixation durations.
Even AUC and NSS could be computed on these three-dimensional predictions.
However, in the case of non-probabilistic models it might not be as obvious how to compute the three-dimensional analogue of the conditional priority maps.

In this work, we quantified how well scanpath models predict human scanpaths under free viewing conditions.
This is also what most included models have been originally build for (for an interesting special case, see, e.g., MASC, which was specifically build to model both free-viewing and visual search).
Free viewing has traditionally received a lot of attention since it was thought to best pronounce involuntary effects such as low-level saliency.
Nevertheless, tasks are known to strongly influence scanpaths at least since the seminal work of \textcite{yarbusEyeMovementsVision1967} and there is a lot research on scanpaths under certain tasks \parencite{rothkopfTaskContextDetermine2007,rothkegelSearchersAdjustTheir2019,yangPredictingGoalDirectedHuman2020}.
Fortunately, nothing in our paradigm is special to free viewing.
It can be directly applied to evaluate how well models predict eye movements under a certain task.
Even evaluating how well models predict eye movements under different tasks can be simply done by extending the conditional fixation predictions to also depend on a task.
This corresponds to asking the model, e.g., ``given that a subject has so far looked at these image locations and given that the subject is looking for a cup, where is the subject most likely to fixate next?''.
In fact, we are already planning to extend our analysis by also including datasets of scanpaths under some tasks to evaluate how well models can predict eye movements in these situations.

The field of computer vision is sometimes criticized for its focus on benchmarking, and, obviously, when trying to understand what drives human scanpaths, benchmarking is not everything.
Nevertheless, our results show that it is important to test proposed mechanisms of fixation selection with respect to their predictive power in a principled way.
For advancing our understanding of scanpaths, a good model should be both interpretable in some way and reach a reasonable benchmark score.
To facilitate future research, we will make the code for our evaluations public.
Beyond that, we will incorporate a scanpath benchmarking track to the MIT/Tuebingen saliency benchmark\footnote{\url{https://saliency.tuebingen.ai}} \parencite{juddBenchmarkComputationalModels2012,kummererSaliencyBenchmarkingMade2018} to allow researchers to score their models in a fair comparison on established benchmarking datasets without the possibility for overfitting.

\paragraph{Acknowledgements:} We thank Lisa Schwetlick and Matthias Tangemann for their helpful comments and feedback on this manuscripts. This work was supported by the German Federal Ministry of Education and Research (BMBF): Tübingen AI Center, FKZ: 01IS18039A and the Deutsche Forschungsgemeinschaft (DFG, German Research Foundation): Germany’s Excellence Strategy – EXC 2064/1 – 390727645 and SFB 1233, Robust Vision: Inference Principles and Neural Mechanisms.



\printbibliography

\clearpage

\section{Supplement}

\subsection{Implementation details of evaluated models}
\label{sec:appendix_methods_models}

\paragraph{Itti-Koch:} 
We use the MATLAB implementation of the model in the SaliencyToolBox \parencite{waltherModelingAttentionSalient2006}\footnote{\url{http://www.saliencytoolbox.net/}}
As conditional saliency map, we use what's called WTA map in the visualizations of the SaliencyToolBox.
More precisely the membrane potential of the leaky-integrate-and-fire-neurons for the input from the saliency map at the state where the network selects a winner.
The original implementation only allows sampling scanpaths.
Since the original implementation only allows to generate scanpaths, we extended it to allow overwriting the sampled fixation positions with given fixation locations as described in Section \ref{sec:modifying_models}.
Additionally, we also evaluate the model as a purely spatial model, using only the computed saliency map without any WTA modulation.

\paragraph{CLE:}
There is a matlab implementation of CLE\footnote{\url{https://www.mathworks.com/matlabcentral/fileexchange/38512}}.
However, this implementation makes use of multiple steps of best-of-k sampling and rejection sampling (see Section \ref{sec:best-of-k}).
Therefore it's not possible to analytically compute the predicted conditional fixation density with this implementation.
Using the results from Section \ref{sec:best-of-k}, we reimplemented the model in python in a way that allows conditioning on a given scanpath history and, instead of sampling, analytically computes the conditional fixation density which then can be sampled from.
CLE requires a saliency map for the viewed image.
The original implementation uses the Itti-Koch model.
We evaluate the model both using saliency maps from the Itt-Koch model and the log density predictions of DeepGaze II.

The model has several parameters that strongly affect the resulting distribution (e.g, temperature, scale of the saliency potential).
Besides evaluating the model using the original parameters, we also optimized the parameters \texttt{k\_P, k\_R=1-k\_P, h, tau\_V, sigma, T} and \texttt{relFoaSize} for maximum likelihood on a subset of 100 random images from the MIT1003, using the L-BFGS-B implementation in scipy \parencite{virtanenSciPyFundamentalAlgorithms2020}.
The number of 100 images was chosen to be a good approximation of the full dataset with respect to the six trained parameters while keeping training times feasible.

As a third version of CLE, we evaluated the model without any saliency information by using constant saliency maps which makes the model effectively an unconstrained Levy flight.

\paragraph{LeMeur16:} We use the original Matlab implementation\footnote{\url{https://people.irisa.fr/Olivier.Le_Meur/publi/2016_VR/index.html}}.
In order to select a next fixation, the model samples multiple candidates from a spatial distribution and chooses the sample with highest probability.
Here, we replace this sampling step by analytically computing the resulting conditional fixation distribution (see Section \ref{sec:best-of-k}). 
We use the model's default parameters except for \texttt{nCandidates}, the number of samples in the last step.
We noticed that using a value of 1 instead of 5 doesn't affect AUC performance but massively increased IG and NSS scores.

\paragraph{MASC:} We use on the original Matlab implementation as provided by the authors and adapt it as detailed in Section \ref{sec:modifying_models}.
MASC computes a conditional priority map in superior colliculus space, where the WTA is applied.
We transform this map back into image space, applying with linear interpolation and nearest neighbor extrapolation to cover all of the image space.
We evaluate the MASC model both using the Itti-Koch model as internal saliency model (as in the original publication) and using the DeepGaze II model.
We removed the blurring operation from DeepGaze II, since MASC applies its own Gaussian blurring in superior colliculus space.

\paragraph{G-Eymol:} We use the original python implementation \footnote{\url{https://gitlab.com/dariozanca/gravitational-eymol}}.
As suggested by the authors, images are rescaled to a maximal size of 224px and the model is applied with the parameters \texttt{fps=60}, \texttt{alpha\_c=0.5}, \texttt{alpha\_of=0.3}, \texttt{dissipation=1.5} and \texttt{max\_distance=224}.
When computing a conditional priority map, we initialize the gaze position at the last fixation position and run the simulation until the fixation detection algorithm detects the next fixation.
\textbf{TODO: mention that we build IoR map over previous fixations excluding initial one}
For the fixation detection, we use \texttt{pygazeanalyser.detectors.fixation\_detection}, as suggested by the authors.
To convert the predicted fixation location into a priority map, we place a Gaussian with a width of 9 degrees of visual angle at predicted fixation location.
The size of the Gaussian has been selected to yield maximal NSS in a grid search on a subset of MIT1003.

\paragraph{STAR-FC:} We adapted the original C++ implementation according to Section \ref{sec:modifying_models}.
For MIT1003 and MIT300, we use the parameters \texttt{pix2deg=35} and \texttt{view\_dist=0.61}, according to the experimental setup of the dataset.
For CAT2000, we use the parameters \texttt{view\_dist=1.06} and \texttt{input\_size\_deg=45.5} (as suggested by the authors for this dataset).

\paragraph{IRL:} We use the original Matlab implementation as provided by the authors and adapted it according to Section \ref{sec:modifying_models}.

\paragraph{Saccadic Flow:} We use the original R implementation as provided by the authors.

\paragraph{SceneWalk:} We use the original Matlab implementation as provided by the authors.
SceneWalk requires an internal saliency model, for which the authors use empirical saliency maps.
In order to make the model image-computable and for fair comparison, here, we instead use density predictions of DeepGaze II.
The model also uses fixation durations.
For the CAT2000 dataset, where we don't have access to fixation durations, we instead randomly sample fixation durations from the MIT1003 dataset.
Obviously this removes some structure from the data but at least it allows to evaluate SceneWalk at all on CAT2000.
Experiments on MIT1003 suggest that randomly sampling fixation durations degrades prediction performance in terms of log likelihood by around 0.01 bit/fixation.
In that case, the model scores should still reflect the model performance compared to other models fairly well.

\paragraph{SaltiNet:} We use the official implementation\footnote{\url{https://github.com/massens/saliency-360salient-2017}}.
The model was originally intended for 360 degree images.
While the model was trained on the iSUN dataset \parencite{xuTurkerGazeCrowdsourcingSaliency2015}, before being finetuned on 360 degree images, the weights for this training are not available.
This means that the model is applied here with a heavy domain shift, so the results should be taken with a grain of salt.
The model predicts a twelve-dimensional saliency volume for an input image.
In order to predict the $n$th fixation of a scanpath, we use the $n$th slice of the saliency volume.
This is a slight difference to the original implementation where the fixations of a scanpath are equally distributed over the saliency volume.
In order to encourage saccades to be not too long, as in the original model, we multiply the prediction with a Gaussian located at the last fixation position and with kernel size $\frac13, \frac23$ relative to image width and height.
The original implementation added this prior only when the last fixation was in the same slice of the saliency volume.
We always apply the prior, which should improve predictions.

\paragraph{DeepGaze II:} We use the published tensorflow model\footnote{\url{https://deepgaze.bethgelab.org/}} which computes predictions for all training crossvalidation folds and then averages them.
When evaluating the MIT1003 dataset, we use for each image the prediction of the crossvalidation fold that had said image in test partition.
For all other datasets we use average of all predictions.
For MIT1003 and the training set of CAT2000 we use the centerbias baseline models (see above) as centerbias.
For MIT300 we use the average prediction of the MIT1003 centerbias model (as done in the original publication).
For the test set of CAT2000, we use the average prediction of the centerbias of the training set.

\paragraph{DeepGaze III:} We use the original implementation of the model as described in \textcite{kummererDeepGazeIIIUsing2019}. Since the model needs two previous fixations as input, we can't apply it to the first fixation after the initial central fixation. For those fixations we use DeepGaze II with a center bias for the first fixation from the fixation number dependent center bias baseline (see above).
For all other fixations we can compute DeepGaze III predictions where we use a centerbias that has been computed in the same way as the overall centerbias, but using only fixations starting from the second voluntary fixation in each scanpath.

\paragraph{IOR-ROI-LSTM:} We use the original implementation of the model\footnote{\url{https://github.com/sunwj/scanpath}}.
We adapted the original sampling code according to Section \ref{sec:modifying_models} to be able to run the model with given scanpath histories instead of sampled fixations.
The model internally predicts for each fixation a mixture of Gaussians.
In order to incorporate the saccade length distribution, it samples multiple (30) candidates from this mixture distribution, computes for each candidate the product of the mixture probability and oculomotor bias term and takes the candidate for which the product is maximally.
If all candidates are outside of the image are, the next fixation is sampled from a center bias.
We apply the results from Section \ref{sec:best-of-k} to compute the closed form conditional fixation density that is the result of this sampling procedure.
As input, the model needs besides the input image a semantic segmentation mask.
As the authors, we use Mask$^{\chi}$-RCNN \parencite{huLearningSegmentEvery2018} via its published code\footnote{\url{https://github.com/ronghanghu/seg_every_thing}} to compute these segmentation masks.
Here, we use the published ResNet50 model with a threshold of 0.4.
To account for occlusion, we overlay the segmentation masks ordered by size of the bounding box.
The model also predicts and uses fixation durations.
For the CAT2000 dataset, where we don't have access to fixation durations, we instead use fixation durations predicted by the model itself.

\subsection{Best-of-k sampling}
\label{sec:best-of-k}

Several existing scanpath models draw multiple candidates from some distribution, compute a gain for each candidate and then select the candidate with the highest gain, e.g., \textcite{boccignoneModellingGazeShift2004}, \textcite{lemeurSaccadicModelEye2015} and \textcite{sunVisualScanpathPrediction2019}.
This pattern, which we will call \textit{best-of-k sampling} in the following, is convenient for sampling fixations but it makes it nontrivial to compute the full conditional model distribution $p(f_i \mid f_0, \dots, f_{i-1}, I)$, which is needed to evaluate model log-likelihoods as we do it here.

Here, we show that the full distribution can actually be computed analytically in this case: 

\paragraph{Theorem}
    Let $X_1, \dots, X_N \sim P$ be $N$ i.i.d. samples of a distribution $P$. Let $f(X)$ be a gain function. Let $Y=\arg\max_{X_i}$ be the random variable that takes the value of the $X_i$ with highest gain $f(X_i)$ out of all $X_1, \dots, X_N$. Then
    \[
      P[Y=y] = \frac{P[X=y]}{P[f(X)=f(y)]}\left(P[f(X) \leq f(y)]^N - P[f(X) < f(y)]^N\right)
    \]

\textit{Proof:} We use the ansatz to first think about the random variable $f(Y)$ or more precisely, CDF of $f(Y)$. The value of $f(Y)$ will be smaller than some $f(y)$ if and only if the gain of all $X_i$ is smaller than $f(y)$:
\begin{align*}
  P[f(Y) \leq f(y)] &= P[f(X_1) \leq f(y), \dots, f(X_N) \leq f(y)]\\
                    &= P[f(X) \leq f(y)]^N
\end{align*}

Therefore we have (in the case of discrete distributions, such as pixel locations):
\[
  P[f(Y) = f(y)] = P[f(X) \leq f(y)]^N - P[f(X) < f(y)]^N
\]
By going back from $f(Y)$ to $Y$, where we get
\[
P[Y=y] = \frac{P[X=y]}{P[f(X)=f(y)]}\left(P[f(X) \leq f(y)]^N - P[f(X) < f(y)]^N\right).
\]

Often candidates can additionally be rejected based on some criteria, e.g., samples outside of the image area.
This can be handled by considering an additional ``image pixel'' which has the probability of a candidate being rejected and is assigned a gain of $-\infty$.

\end{document}